\documentclass[acmlarge,nonacm]{acmart}
\AtBeginDocument{%
  }

\usepackage{natbib}
\usepackage{float}

\begin{document}

\title{TunnElQNN: A Hybrid Quantum-classical Neural Network for Efficient Learning
 }
\author{A. H. Abbas}
\email{aborae@csu.edu.au}
\orcid{1234-5678-9012}
\authornotemark[1]
\email{aborae@csu.edu.au}
\affiliation{%
  \institution{Artificial Intelligence and Cyber Futures Institute,
Charles Sturt University, Bathurst, NSW, 2800, Australia}
  \city{Bathurst}
  \state{ NSW, 2800}
  \country{Australia}
}

\setcopyright{none} 

\renewcommand{\shortauthors}{A. H. Abbas et al.}

\keywords{Hybrid Quantum-classical machine learning model, quantum circuit, and neural network}

\maketitle
\section{Abstract}
Hybrid quantum-classical neural networks (HQCNNs) represent a promising frontier in machine learning, leveraging the complementary strengths of both models. In this work, we propose the development of TunnElQNN, a non-sequential architecture composed of alternating classical and quantum layers. Within the classical component, we employ the Tunnelling Diode Activation Function (TDAF), inspired by the I-V characteristics of quantum tunnelling. We evaluate the performance of this hybrid model on a synthetic dataset of interleaving half-circle for multi-class classification tasks with varying degrees of class overlap. The model is compared against a baseline hybrid architecture that uses the conventional ReLU activation function (ReLUQNN). Our results show that the TunnElQNN model consistently outperforms the ReLUQNN counterpart. Furthermore, we analyse the decision boundaries generated by TunnElQNN under different levels of class overlap and compare them to those produced by a neural network implementing TDAF within a fully classical architecture. These findings highlight the potential of integrating physics-inspired activation functions with quantum components to enhance the expressiveness and robustness of hybrid quantum-classical machine learning architectures.

\section{Introduction}
Classical neural networks (CNNs) have achieved remarkable success, yet they face limitations such as vanishing gradients, redundancy, and inefficiency in high-dimensional tasks. Quantum machine learning (QML) offers a compelling alternative by leveraging quantum principles such as entanglement, superposition, and quantum parallelism. These principles enable powerful computational strategies that improve information processing \cite{nielsen10}. However, practical implementation of quantum algorithms is currently hindered by the limitations of Noisy Intermediate-Scale Quantum (NISQ) devices, including short qubit coherence times, gate errors, and noise \cite{preskill18}.

Despite advancements, scalability and reliance on NISQ hardware remain major challenges. Models utilising more than four qubits often suffer from noise-induced degradation \cite{zaman24}, and real-world deployment is limited by hardware constraints. For instance, variational quantum circuit (VQC)-based neurons demonstrate a 10\% increase in simulated accuracy but perform poorly on physical NISQ devices \cite{arthur23}. While techniques such as deep residual learning help improve robustness in noisy environments \cite{mari21}, error-resilient training protocols and multi-class extensions remain active areas of research \cite{hqpin25}.

Hybrid quantum-classical neural networks (HQCNNs) provide a pragmatic path forward by leveraging classical computational power alongside quantum-enhanced feature extraction~\cite{schuld19, cerezo21}. They address this challenge by integrating quantum circuits with classical architectures, leveraging the high-dimensional Hilbert space of quantum systems to enhance learning capacity and efficiency, while remaining compatible with current hardware \cite{schuld16}. HQCNNs exploit quantum capabilities to enhance learning expressiveness, particularly in tasks involving complex data and non-linear patterns. Classical layers handle high-dimensional data and conventional tasks, whereas quantum layers exploit qubit entanglement and superposition to capture complex correlations more efficiently \cite{ornl23,biamonte17}.

Early work by Schuld et al. demonstrated that quantum circuits embedded in classical frameworks could generate quantum states efficiently through quantum parallelism \cite{schuld16}. Subsequent advances, such as VQNet, integrated variational quantum circuits into classical training loops \cite{vqe2019}, while architectures such as quantum Convolutional neural networks (QCCNNs) \cite{cong19} and quantum residual networks \cite{residual2021} showcased applications in image classification and optimisation \cite{Amira21, mari21}. Recent studies highlight HQCNNs outperforming classical models in constrained data regimes or high-dimensional tasks \cite{qclassify2022}, with applications that span quantum-enhanced transfer learning \cite{qtransfer2024} and communication systems \cite{beamforming2024}. It has also been reported that HQCNNs achieve improvements in classification accuracy and better cost minimisation compared to standalone quantum models \cite{ornl23}. Additionally, combining quantum and classical neural networks can significantly reduce the number of training parameters while enhancing model accuracy \cite{Liu24}.

HQCNNs demonstrate versatility across fields. In drug discovery, quantum-enhanced feature maps accelerate molecular property prediction \cite{saki23}, while hybrid models in quantum chemistry optimise molecular simulations with fewer parameters than classical counterparts \cite{saki23}. Image recognition benefits from QCCNNs, which improve accuracy by 10–15\% on NISQ devices compared to CNNs \cite{Amira21}. Compact HQCNNs, such as the two-qubit H-QNN model, achieve 90.1\% accuracy in binary image classification by integrating quantum circuits with classical Convolutional backbones, outperforming CNNs (88.2\%) and showcasing robustness against overfitting \cite{blaszczyk23}. 

Physics has increasingly influenced machine learning, inspiring algorithms that exploit physical laws to improve computational and energy efficiency \cite{Mar20, Nak22, Mak23_review, Abb24_1, maksymov24}. Physical reservoir computing, for instance, uses the natural dynamics of physical systems for tasks like time series prediction and pattern recognition, offering energy-efficient alternatives to conventional models \cite{Mar20_2, Dud23, Abb24, Abb25, abdelghani25}. 

Extending the integration of physical principles into machine learning, we have recently introduced a CNN that employs the current-voltage (I-V) characteristic of a tunnel diode as a novel, physics-based activation function. The activation function is a central component in classical models, whose nonlinearity is essential for capturing complex input-output relationships and enabling generalisation. Without it, neural networks become limited linear models. This tunnel-diode activation function (TDAF) surpasses traditional functions such as ReLU in both accuracy and loss, and its compatibility with electronic circuit implementation opens new possibilities for neuromorphic and quantum-inspired AI hardware \cite{McNaughton25}. Such designs are particularly valuable in environments where qubit-based quantum computing remains impractical, offering a practical path toward scalable, energy-efficient AI systems.

In this study, we introduce TunnELQNN, a non-sequential hybrid architecture composed of alternating classical and quantum layers. In the classical component, we employ the TDAF as the nonlinearity. The quantum component consists of a 2-qubit circuit arranged in a non-sequential configuration with angle embedding and entangling gates, enabling the hybrid model to capture complex, nonlinear relationships that are often challenging for purely classical models. To evaluate the performance of this hybrid model, we use a synthetic dataset of interleaving half-circle designed for multi-class classification tasks with varying degrees of class overlap.

We compare TunnELQNN against a baseline hybrid model that uses the conventional ReLU activation function (ReLUQNN). Our results demonstrate that TunnELQNN consistently outperforms the ReLUQNN counterpart. In addition, we analyse the decision boundaries formed by TunnElQNN under different levels of class overlap and compare them to those produced by a fully classical neural network implementing TDAF. These findings underscore the potential of integrating physics-inspired activation functions with quantum components to enhance the expressiveness and robustness of hybrid quantum-classical machine learning systems.

The remainder of this paper is organised as follows. In Section 2, we present the model and describe the network architecture. In Section 3, we compare the performance of TunnELQNN against a baseline hybrid model, ReLUQNN. We also analyse the decision boundaries generated by TunnELQNN under varying degrees of class overlap and compare them with those produced by a fully classical neural network utilising TDAF. Finally, we conclude with a summary of the findings and provide recommendations for future work.
\begin{figure}[b]
    \centering
    \includegraphics[width=\linewidth]{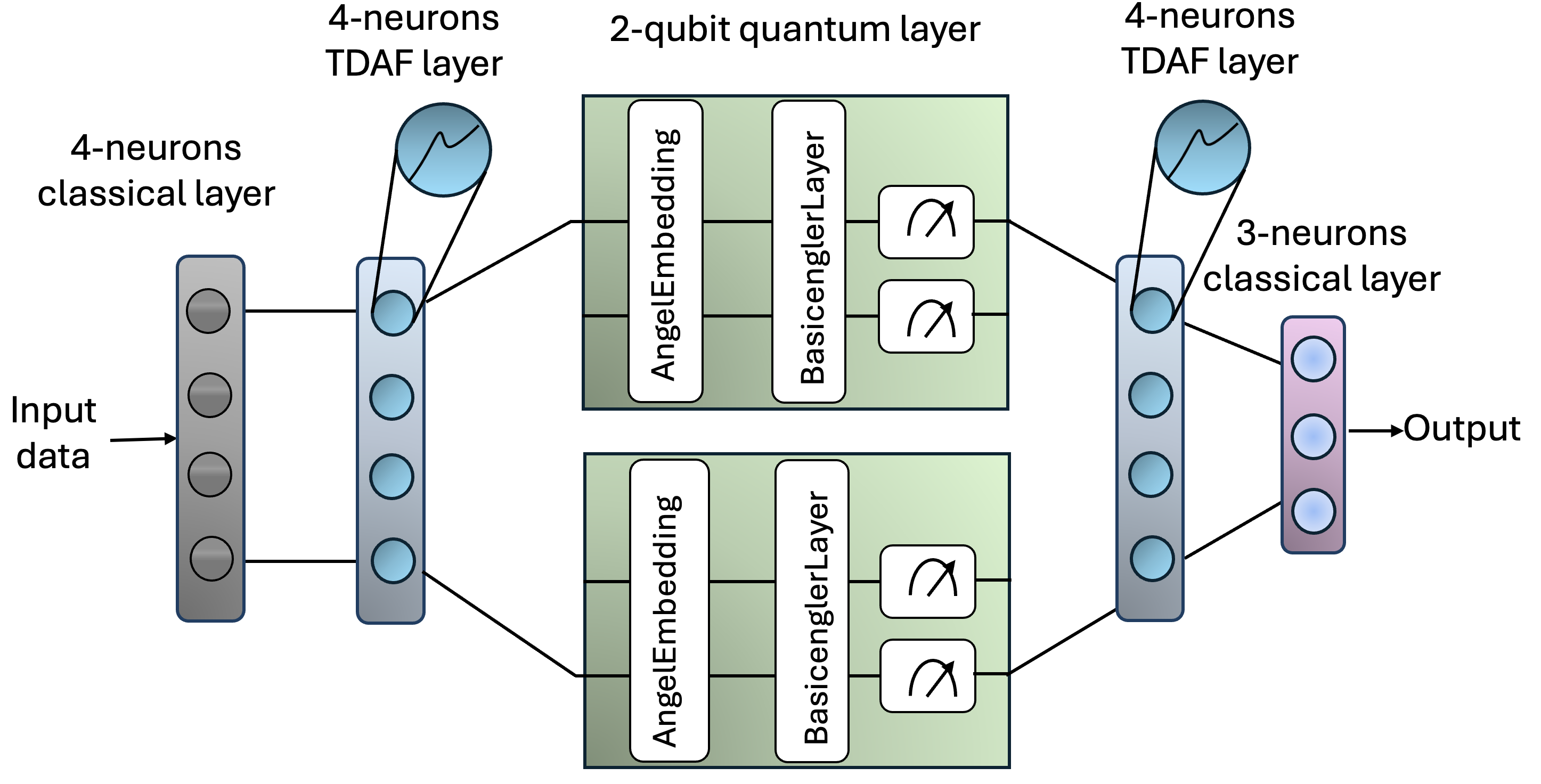}
    \caption{Hybrid quantum-classical neural network architecture combining classical layers with Tunnelling-Diode Activation Function (TDAF) units and a 2-qubit quantum layer featuring AngleEmbedding and entangling operations. The model processes input data through alternating classical and quantum layers to produce a 3-neuron output, leveraging both classical and quantum computational advantages.}
    \label{fig1}
\end{figure}
\section{TunnELQNN Architecture: Quantum-Classical Hybrid Design}
The hybrid architecture depicted in Figure.~\ref{fig1} integrates classical and quantum computational layers within a framework designed to process a complex synthetic dataset comprising interleaved half-circles for multi-class classification tasks with varying degrees of class overlap.

The computational flow begins with a classical input layer—a linear transformation comprising four neurons, which captures the input features. These features are subsequently transmitted to a TDAF layer \cite{McNaughton25}, also composed of four neurons. This layer introduces essential non-linearity into the system, enhancing its capacity to model complex patterns. The output from the TDAF layer is then forwarded to a non-sequential quantum layer comprising two qubits. Within this quantum section, \textit{AngleEmbedding} \cite{angleembedding} is employed to encode the classical inputs into quantum states by mapping feature values to the rotational angles of quantum gates. 

The embedded qubits are then processed via a \textit{BasicEntanglerLayer} \cite{basicentanglerlayers}, which applies entangling operations to establish correlations between qubits—enabling the representation of higher-order joint features. Quantum measurements are then performed on each qubit, yielding classical outputs from the entangled quantum states.

The resulting quantum-derived outputs are subsequently passed through an additional TDAF layer, again with four neurons, to ensure a smooth computational transition between quantum and classical regimes. This is followed by a classical processing layer with three neurons, serving as the penultimate stage before producing the final output of the network.

This architecture leverages the strengths of both classical and quantum computation. Classical layers offer efficient data representation and initial processing, while quantum layers contribute enhanced feature encoding and entanglement-based correlations. This quantum-classical interface facilitate the parallel information processing across subspaces. Such a design makes this architecture particularly suitable for tasks involving classification tasks, and pattern recognition.

\subsection{Tunnelling Diode Activation Function (TDAF): A Physics-Inspired Nonlinearity}
The current-voltage (I–V) response of a tunnel diode exhibits strong nonlinearity, primarily due to the presence of a region with negative differential resistance \cite{McNaughton25}. This behaviour can be described by the expression:
\begin{equation} \label{eq:1}
    I(V) = J_{1}(V)+J_{2}(V)\,,
\end{equation}
where the individual components are defined as
\begin{eqnarray} \label{eq:term1}
    J_{1}(V) &=& a\ln \left( \frac{1 + e^{\alpha + \eta V}}{1 + e^{\alpha-\eta V}} \right)  \times \left(\frac{\pi}{2} + \tan^{-1} \left( \frac{c - n_1V}{d} \right)\right)\, \nonumber\\
    J_{2}(V) &=& h \left(e^{\gamma V} - 1 \right)
\end{eqnarray}

The parameters are given by $\alpha = \frac{q (b - c)}{k_B T}$, $\eta = \frac{q n_1}{k_B T}$, and $\gamma = \frac{q n_2}{k_B T}$, where $q$ denotes the electron charge, $k_B$ is Boltzmann’s constant, $T$ is the temperature, $V$ is the voltage, and $a$, $b$, $c$, $d$, $n_1$, and $n_2$ are system-specific parameters. In this study, we adopt the values $T = 300$,K, $a = 0.0039$,A, $b = 0.5$,V, $c = 0.0874$,V, $d = 0.0073$,V, $n_{1} = 0.0352$, $n_{2} = 0.0031$, and $h = 0.0367$,A, consistent with Ref.~\cite{OPiwonka21}, except for $b$, which has been increased by a factor of $10$ to investigate the nonlinear dynamics of TDAF (for further details, see Ref.~\cite{McNaughton25}).

\begin{figure}[h]
    \centering
    \includegraphics[width=\linewidth]{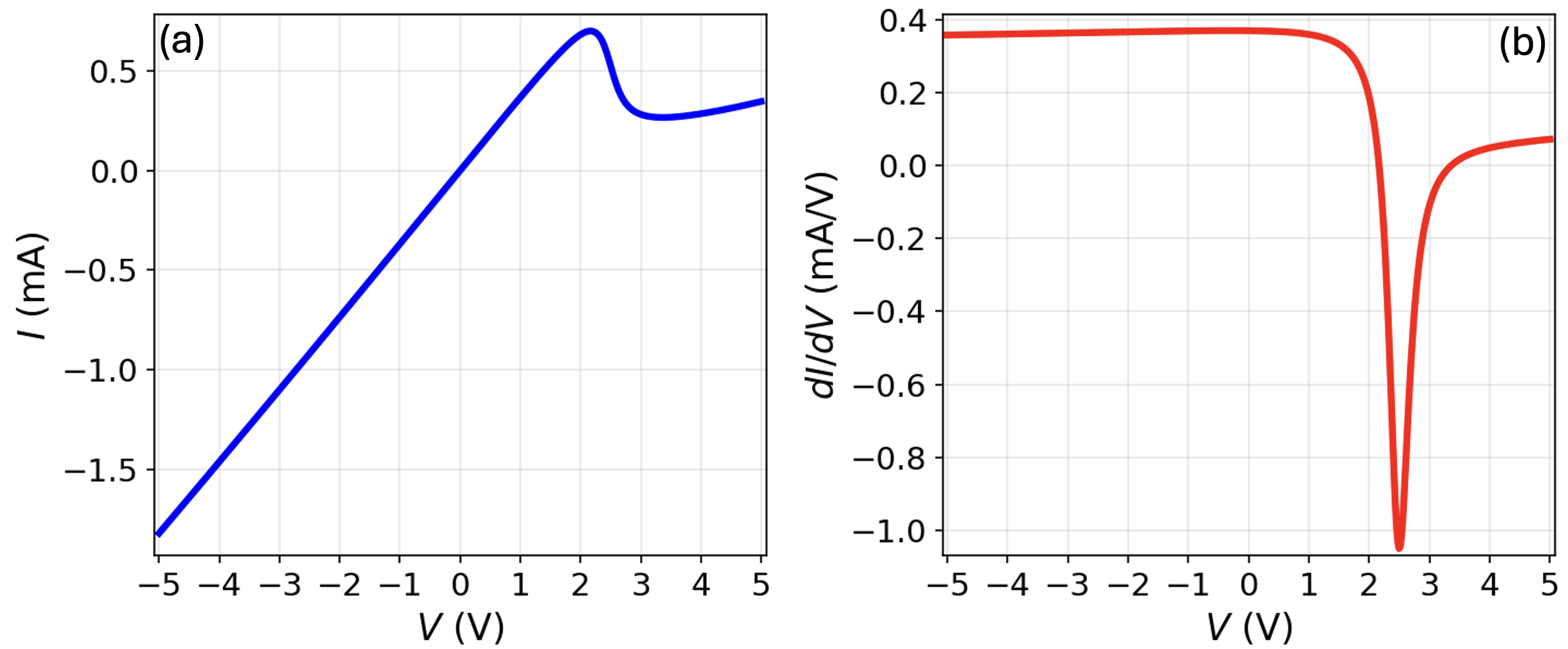}
    \caption{(a) The I–V response of the tunnel diode. (b) Its corresponding differential conductance. }
    \label{fig2}
\end{figure}

The I–V characteristic of the TDAF defined in Eq.~(\ref{eq:1}) comprises three distinct regions: two segments where the differential resistance is positive, separated by a central region exhibiting negative differential resistance (approximately within the interval \( V \in [1, 3] \), see Fig.~\ref{fig2}). The term \( J_{1}(V) \) governs the low-voltage behaviour, producing both the initial positive differential resistance and the NDR region. However, it does not capture the resurgence of current at higher voltages. The term \( J_{2}(V) \), dominant at elevated voltages, accounts for this second region of positive differential resistance~\cite{Sch96}. 

\subsection{Non-sequential Quantum Layer: Angle Embedding and Entanglement}
Mapping classical data to quantum states is a fundamental step in HQCNNs and significantly impacts their learning capabilities. Several strategies have been proposed to effectively encode classical information into quantum systems. Among these, the two most commonly employed methods in HQCNNs are \textit{AmplitudeEmbedding} and \textit{AngleEmbedding}.

In this work, we adopt \textit{AngleEmbedding}, where each classical feature is mapped to the rotation angles of quantum gates (RX, RY, and RZ). Specifically, each feature governs the rotation of a single qubit around the X, Y, and Z axes. As a result, encoding \(n\) features requires \(n\) qubits, with each qubit representing one feature. This method leverages the quantum system’s ability to explore a high-dimensional Hilbert space, exploiting superposition and entanglement to potentially capture complex data structures beyond classical models. Once the features are encoded, the quantum circuit processes the resulting quantum states through entangling operations and measurements to extract key features for classification. Mathematically, the \textit{AngleEmbedding} can be expressed as:

\begin{equation}
    U{x_i} = \prod_{\rho=X,Y,Z} R_\rho(x_i), \quad \text{where } R_\rho(x_i) := \exp(-ix_i \rho)
    \label{eq:angle_encoding}
\end{equation}

The overall quantum state after applying the rotations is:

\[
|\psi(x_i)\rangle =U{x_i}|0\rangle^{\otimes n},
\]
where \( |0\rangle^{\otimes n} \) denotes the initial state of \(n\) qubits, and the product indicates the sequential application of rotations to each qubit. Each qubit undergoes rotations determined by its respective input feature \(x_i\).

After feature encoding, the quantum circuit is constructed with layers of single-qubit rotations followed by entanglement operations. Entanglement — a uniquely quantum feature — enables qubits to share information non-locally, allowing the system to model complex, non-linear relationships inaccessible to classical networks. In our architecture, each quantum layer comprises single-qubit rotations parameterised by trainable angles, followed by a ring of CNOT gates to establish entanglement. In the CNOT ring, each qubit \(q_i\) is entangled with its neighbour \(q_{i+1}\), and the last qubit is connected back to the first \cite{basicentanglerlayers}:

\[
BasicEntanglerLayers: \prod_{i=1}^{n} \text{CNOT}(q_i, q_{i+1})
\]
This closed-ring topology ensures that all qubits are interconnected, promoting global quantum correlations across the system. Without connecting the first and last qubits, entanglement would be restricted to neighbouring pairs, thereby limiting the model's expressiveness.

Our full quantum circuit architecture consists of two non-sequential core blocks, with the option to repeat this block multiple times to deepen the quantum transformations. Finally, we perform measurements in the Pauli-Z basis, yielding classical outputs in the form of expectation values for each qubit. These quantum-processed features are then passed to a classical neural network for the final classification step.

\section{Model Training Protocol and Dataset Description}
The model is trained on a three-class interleaving half-circles dataset using the Adam optimiser (learning rate = 0.02) and cross-entropy loss to monitor performance. Training is conducted over 150 epochs with a batch size of 128, tracking both loss and accuracy.

\begin{figure}[h]
    \centering
    \includegraphics[width=0.75\linewidth]{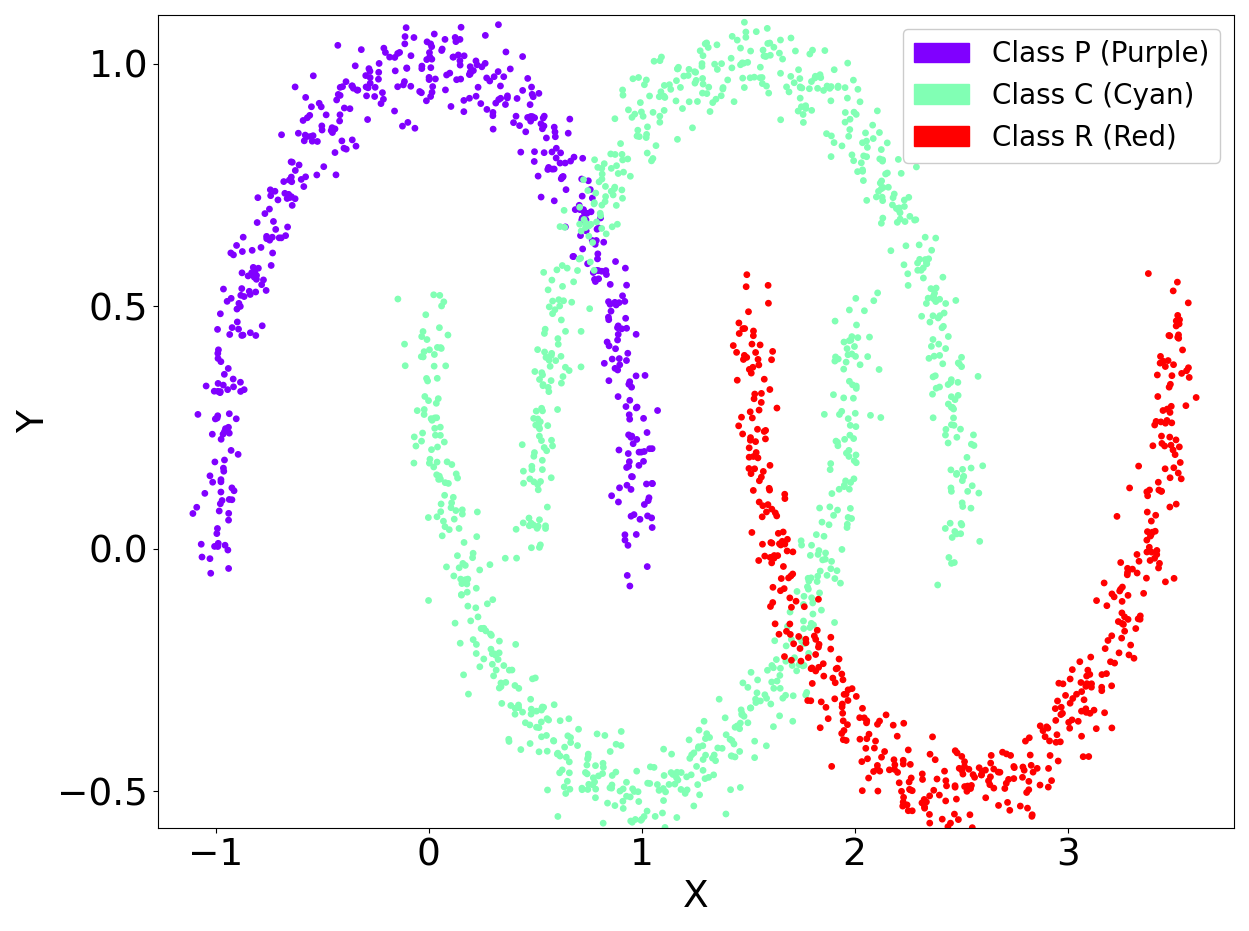}
    \caption{Three interleaving half-circles dataset with $n=2000$ samples. Data points are coloured by class: Class P for Purple , Class C for Cyan, and Class R for Red. A horizontal shift of 1.5 arb. Units was applied to separate the respective half-circles.}
    \label{fig3}
\end{figure}

Figure \ref{fig3} illustrates the dataset, which consists of three classes: Class P (Purple), Class C (Cyan), and Class R (Red), with a horizontal shift of 1.5 arb. Units. This dataset is used to evaluate the performance of TunnElQNN with three quantum layers against a baseline ReLUQNN with the same architecture. Additionally, the performance of TunnElQNN is assessed under varying horizontal shifts and compared to a standalone CNN using TDAF as the activation function.
\section{Experimental Results and Performance Evaluation}
\subsection{Comparative Performance: TunnElQNN vs. ReLU-based Hybrid Models(ReLUQNN)}
In our implementation, we used a synthetic 2D three-class interleaving half-circles dataset, shown in Fig.\ref{fig3}, with two features per sample. The dataset was split into 80\% for training and 20\% for testing.

\begin{figure}[h]
    \centering
    \includegraphics[width=\linewidth]{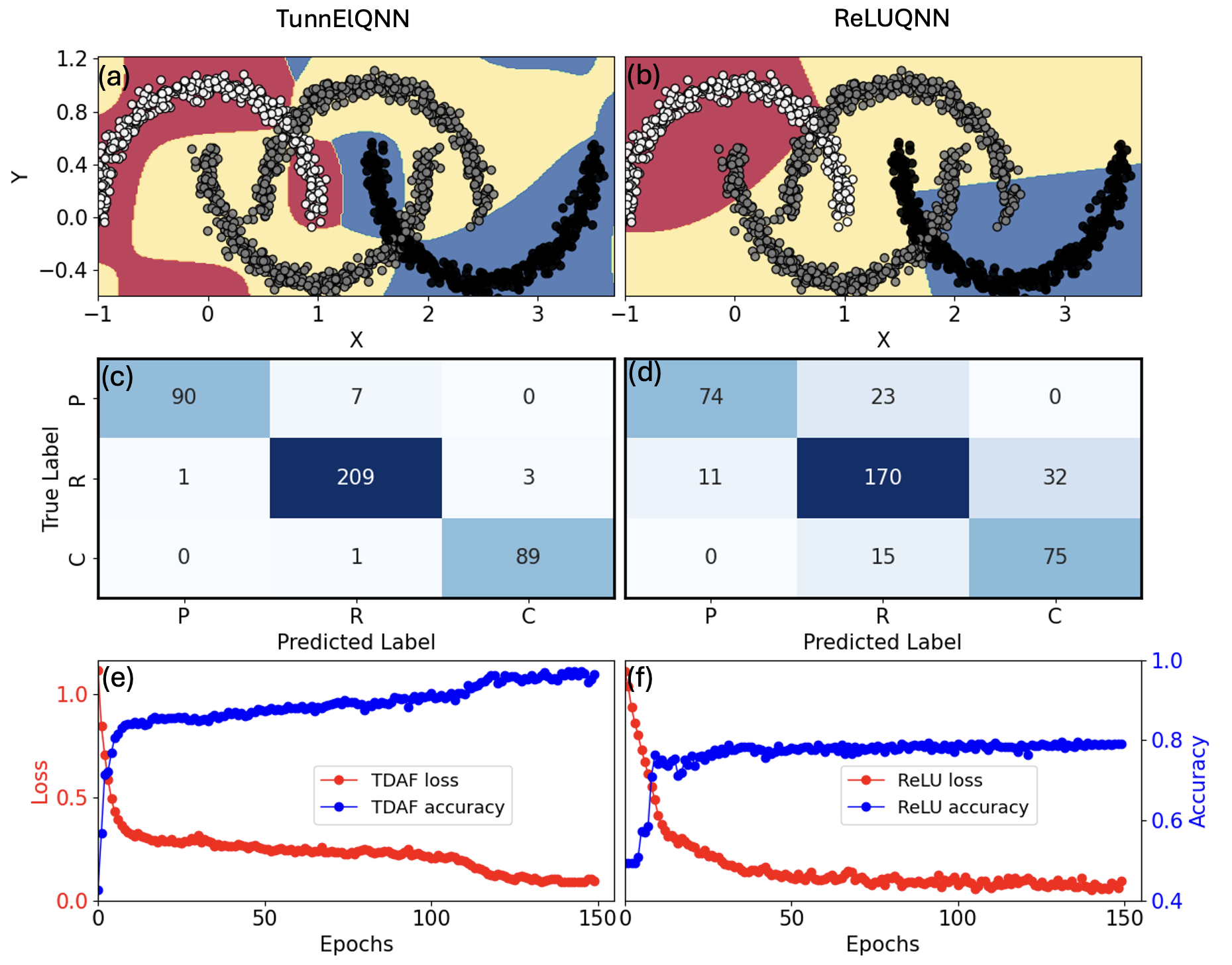}
    \caption{Comparison of TunnElQNN and ReLUQNN on the synthetic 2D three-class interleaving half-circles dataset. The first row illustrates the decision boundaries learned by each model (a: TunnElQNN, b: ReLUQNN). The second row shows the confusion matrices, highlighting classification performance (c: TunnElQNN, d: ReLUQNN). The third row presents training accuracy and loss curves, demonstrating convergence behaviour (e: TunnElQNN, f: ReLUQNN). TunnElQNN exhibits smoother decision regions and achieves 99\% training accuracy, indicating better generalisation, while ReLUQNN reaches 80\% accuracy. achieves higher classification accuracy with minimal misclassification across classes P, R, and C. Training curves showing loss (red, left axis) and accuracy (blue, right axis) over 150 epochs for TunnElQNN (e) and ReLUQNN (f) models.}
    \label{fig4}
\end{figure}

Figure.~\ref{fig4} compares the performance of the TunnElQNN and ReLUQNN models on this classification task. The top panels (a–b) illustrate the decision boundaries for each model. TunnElQNN produces well-defined, class-consistent regions with minimal overlap (Fig.\ref{fig4}(a)), while ReLUQNN yields more fragmented boundaries and greater misclassification, particularly in central overlapping areas Fig.\ref{fig4}(b). The middle panels (c–d) show the confusion matrices for each model. TunnElQNN achieves near-perfect classification with high accuracy across all classes (Fig.\ref{fig4}(c)), whereas ReLUQNN exhibits greater confusion between classes (Fig.\ref{fig4}(d)).

The bottom panels (e–f) display training loss and accuracy over 150 epochs. TunnElQNN converges rapidly, reaching over 99\% accuracy by epoch 130 with steadily decreasing loss Fig.\ref{fig4}(e). In contrast, ReLUQNN converges more slowly, stabilising at 80\% accuracy and retaining higher loss throughout Fig.~\ref{fig4}(f). Validation accuracy also differs significantly: TunnElQNN achieves 97\%, outperforming ReLUQNN’s 87\%. These training dynamics, as shown in panels (e–f), demonstrate the ability of the TunnElQNN model to converge quickly and stably, with minimal residual loss, compared to the slower and less effective learning of the ReLU model.

These results underscore the advantage of trainable, data-adaptive activation functions like TDAF in hybrid quantum-classical systems. TunnElQNN, equipped with TDAF, not only fits the training data more effectively but also generalises better to unseen examples, making it a strong candidate for complex pattern recognition tasks.

\subsection{Evaluating TunnElQNN performance Against Classical TDAF Networks}
In this section, we evaluate and compare the performance of TunnElQNN and a standalone CNN that utilises TDAF as an activation function on the synthetic 2D three-class interleaving half-circles dataset. The dataset features varying degrees of class overlap, which can be achieved by adjusting the horizontal shift. This task tests the ability of the model to recognise boundaries between classes under different levels of separation.
\begin{figure}[H]
    \centering
    \includegraphics[width=0.95\linewidth]{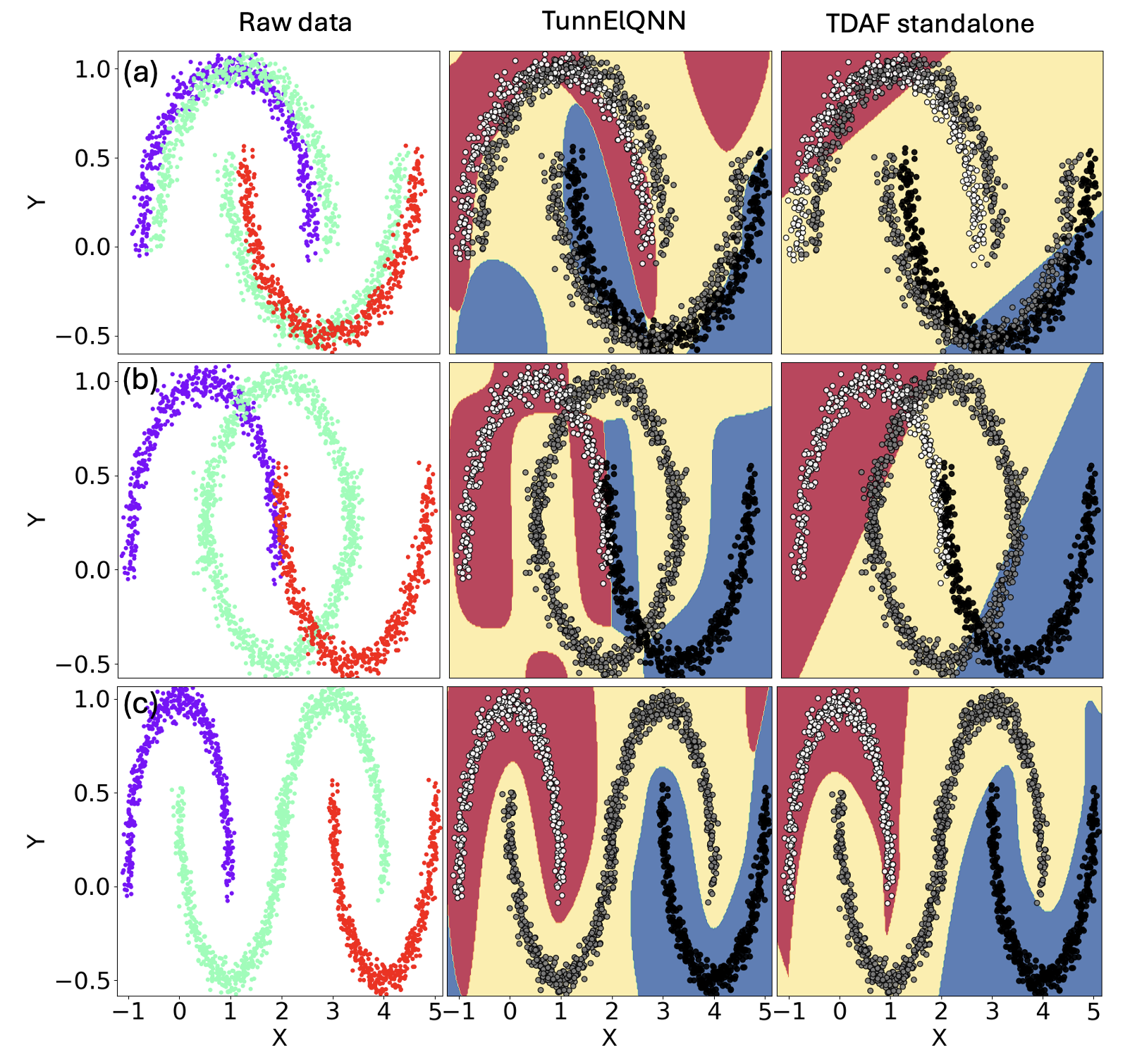}
    \caption{Comparison of the performance of TunnElQNN and a classical neural network using TDAF on the synthetic 2D three-class interleaving half-circles dataset with varying class horizontal separation. Decision boundaries are shown for increasing levels of separation: (a) separation = 0.2 arb. Units (high overlap), (b) separation = 1 arb. Units (moderate overlap), and (c) separation = 3 arb. Units (well-separated classes). TunnElQNN consistently learns accurate decision boundaries across all levels of overlap, while the standalone TDAF-based network performs reliably only when the data is well separated.}
    \label{fig5}
\end{figure}
To begin, we examine how each model performs when the class overlap changes from high to moderate to low, represented by varying degrees of class separation. Specifically, we consider three scenarios for class separation: (a) high overlap (separation = 0.2 arb. Units), (b) moderate overlap (separation = 1 arb. Units), and (c) low overlap (separation = 3 arb. Units), which allows us to assess the robustness of the models under different conditions.

As illustrated in Figure~\ref{fig5}, TunnElQNN with three quantum layers consistently learns accurate decision boundaries across all levels of overlap, maintaining reliable performance even when the class separation is minimal. In contrast, the standalone CNN with TDAF demonstrates high reliability only when the data is well-separated (i.e., scenario (c)). For the high and moderate overlap scenarios (i.e., (a) and (b)), the CNN struggles to maintain accurate decision boundaries, highlighting its limitations in handling complex, overlapping data distributions.

This comparison highlights the advantage of incorporating quantum layers in the TunnElQNN model, particularly in handling datasets with varying degrees of overlap. TunnElQNN demonstrates superior generalisation and robustness across noise levels, while the TDAF-based CNN, though effective in more distinct classification tasks, is less capable when facing closely entangled classes.
\begin{figure}[h]
    \centering
    \includegraphics[width=0.7\linewidth]{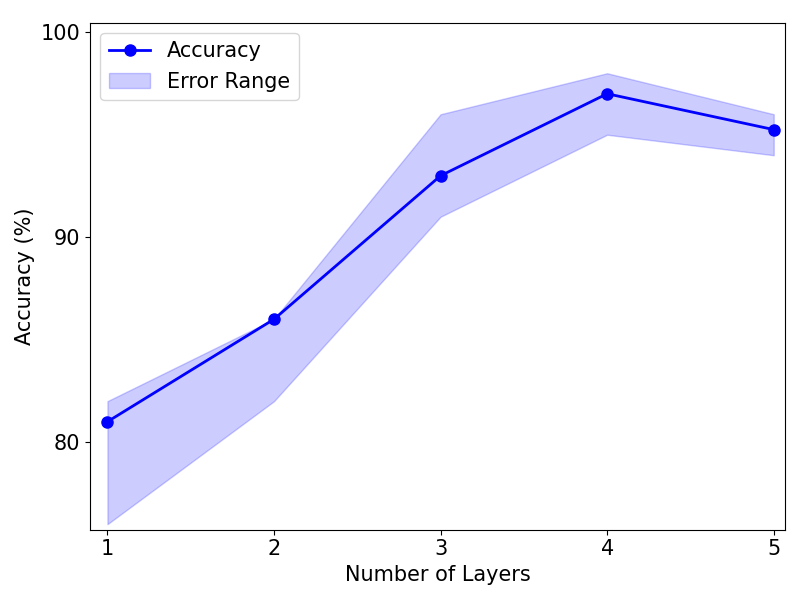}
    \caption{Classification accuracy of the TunnElQNN model as a function of the number of quantum layers. The model was evaluated using the raw dataset shown in Fig.~\ref{fig5}(b). Accuracy improves with increased layer depth, reaching optimal performance at four layers. Beyond this point, performance slightly declines, suggesting diminishing returns with deeper quantum circuits. The shaded region represents the error range across trials. }
    \label{fig6}
\end{figure}

For completeness, we evaluate how the classification accuracy of TunnElQNN varies with the depth of its quantum layer (i.e., the number of quantum layers). To ensure consistency, we conduct experiments using the raw dataset shown in Fig.~\ref{fig5}(b), varying the number of quantum layers in the model. 

The results illustrated in Figure~\ref{fig6} demonstrate the impact of quantum layer depth on the classification accuracy of the TunnElQNN model. As the number of quantum layers increases from one to four, the accuracy steadily improves, indicating that deeper quantum circuits provide enhanced representational power and learning capacity. The performance peaks at four layers, suggesting that this depth achieves a favourable balance between model complexity and effective learning. However, beyond this point, a slight decline in accuracy is observed. 

This degradation may be attributed to overfitting or vanishing gradient problem arising from the deeper quantum architecture. Moreover, deeper circuits are more susceptible to parameter entanglement and vanishing gradients, which can hinder convergence and reduce generalisation performance.

\section{Discussion}
The performance of TunnElQNN highlights the synergy between quantum processing and physics-inspired activation functions. The TDAF, rooted in the nonlinear I–V characteristics of tunnel diodes, offers rich gradient dynamics and avoids saturation effects in traditional activation functions \cite{McNaughton25}. The empirical results demonstrate that this integration yields smoother decision boundaries, better classification accuracy, and improved generalisation—even under high class overlap—compared to both ReLU-based HQCNNs and standalone classical networks using TDAF.

Nevertheless, the model introduces trade-offs. For instance, training with TDAF may increase the computational time especially for deeper model. This overhead is attributable to the complexity of evaluating the nonlinear function. Additionally, deeper quantum circuits (beyond 4 layers) slightly degrade performance, likely due to increased entanglement complexity and gradient vanishing, which remains a known challenge in deep quantum architectures. These results highlight the value of integrating quantum entanglement and trainable, physics-informed activations in hybrid systems. 
\section{Conclusion and Future Directions}
This study introduces TunnElQNN, a non-sequential hybrid quantum-classical neural network leveraging the TDAF to augment classical and quantum learning capacity. Experimental evaluation on a synthetic classification task shows that TunnElQNN significantly outperforms conventional hybrid models (e.g., ReLUQNN) in both accuracy and robustness across varied class overlaps.

To further advance the capabilities and broaden the applicability of the TunnElQNN architecture, several research directions should be pursued. First, testing the model on real NISQ devices is essential to evaluate its resilience to hardware-induced noise and to validate simulation-based findings under practical constraints. Additionally, exploring implementation on analog quantum co-processors could leverage the electronic compatibility of the TDAF, potentially reducing latency and bypassing digital-to-analog conversion bottlenecks. Expanding the model's application to higher-dimensional, real-world datasets—such as medical imaging or time series forecasting—would help assess its generalisation capabilities in complex domains. Finally, increasing the number of quantum layers beyond four, supported by error mitigation techniques or gradient-preserving architectures, may unlock further improvements in model expressivity while managing the known challenges of deep quantum circuits.

\section{Acknowledgement}
The author would like to thank Ivan Maksymov for the valuable discussions and insightful feedback provided during the preparation of this manuscript.
\bibliographystyle{unsrtnat}
\bibliography{sample-base}

\appendix

\end{document}